\documentclass[10pt,twocolumn,letterpaper]{article}

\usepackage{iccv}
\usepackage{times}
\usepackage{epsfig}
\usepackage{graphicx}
\usepackage{amsmath}
\usepackage{amssymb}
\usepackage{rotating}
\usepackage{booktabs}
\usepackage{multirow}
\usepackage{multicol}
\usepackage{subcaption}
\RequirePackage[format=plain,labelformat=simple,labelsep=period,font=small,compatibility=false]{caption}
\RequirePackage[font=footnotesize,skip=3pt,subrefformat=parens]{subcaption}
\usepackage[pagebackref=true,breaklinks=true,letterpaper=true,colorlinks,bookmarks=false]{hyperref}

\iccvfinalcopy 
\ificcvfinal\fi

\begin{document}

\title{Normalizing Flows for Human Pose Anomaly Detection}

\author{
Or Hirschorn\\
Tel-Aviv University, Israel\\
{\tt\small or.hirschorn@gmail.com}
\and
Shai Avidan\\
Tel-Aviv University, Israel\\
{\tt\small avidan@eng.tau.ac.il}
}

\maketitle
\ificcvfinal\thispagestyle{empty}\fi

\begin{abstract}
Video anomaly detection is an ill-posed problem because it relies on many parameters such as appearance, pose, camera angle, background, and more.
We distill the problem to anomaly detection of human pose, thus decreasing the risk of nuisance parameters such as appearance affecting the result. Focusing on pose alone also has the side benefit of reducing bias against distinct minority groups.

Our model works directly on human pose graph sequences and is exceptionally lightweight ($\sim$1K parameters), capable of running on any machine able to run the pose estimation with negligible additional resources.
We leverage the highly compact pose representation in a normalizing flows framework, which we extend to tackle the unique characteristics of spatio-temporal pose data and show its advantages in this use case.

The algorithm is quite general and can handle training data of only normal examples as well as a supervised setting that consists of labeled normal and abnormal examples.

We report state-of-the-art results on two anomaly detection benchmarks - the unsupervised ShanghaiTech dataset and the recent supervised UBnormal dataset.
Code available at~~~\url{https://github.com/orhir/STG-NF}.
\end{abstract}

\section{Introduction}
\begin{figure*}
\centering
\includegraphics{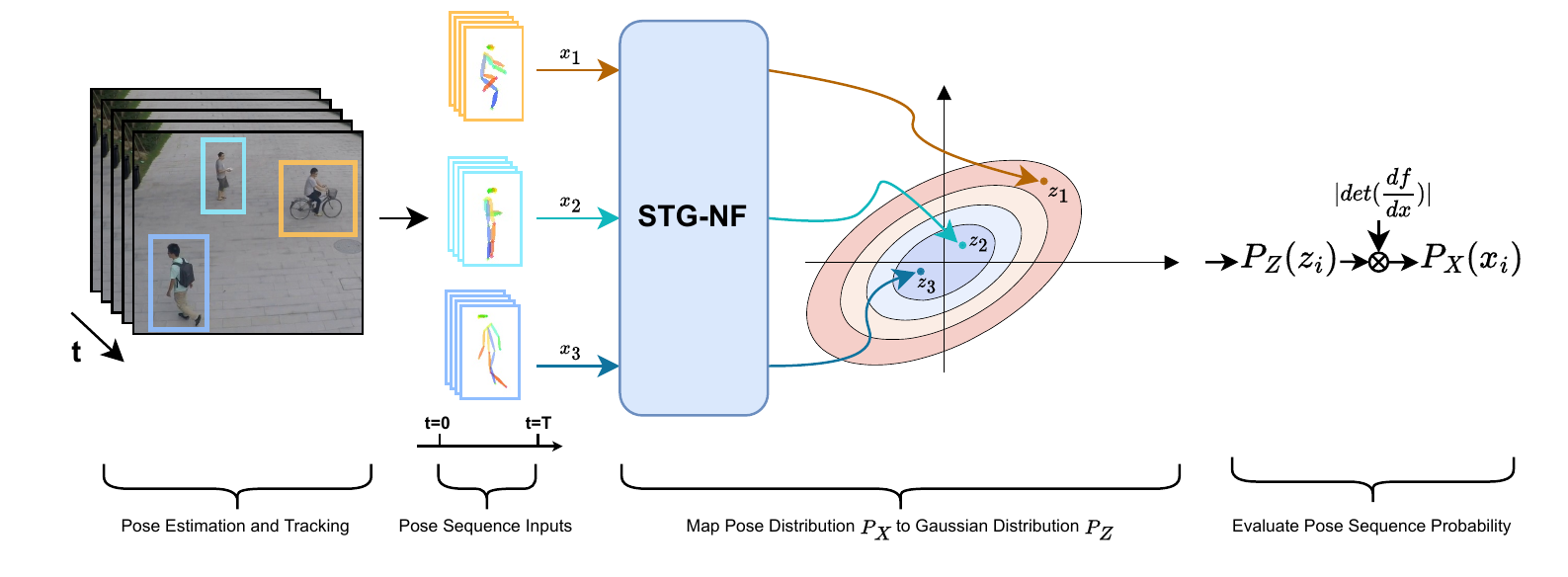}
\caption{\textbf{Framework Overview:} An overview of our unsupervised framework. We perform pose estimation and tracking on the input video. Our Spatio-Temporal Graph Normalizing Flows (STG-NF) network processes each pose sequence separately. During training, our model learns a bijective mapping from data distribution $P_X$ (pose sequences) to a latent Gaussian distribution $P_Z$. Training is done by minimizing the negative log-likelihood of the training data, leveraging the invertibility of our architecture, and using the change of variables formula. During inference, we estimate the probability of each pose sequence.}
\label{fig:arch}
\end{figure*}

 Research on human anomaly detection aims to detect unusual events such as violence or people needing assistance due to an accident~\cite{WhereAreWeWithHumanPoseEstimation}. As these instances are time-critical, the surveillance footage requires immediate processing to enable instant support on-site. This raises the need for fast and reliable methods to detect abnormal events.

Although investigated extensively~\cite{Acsintoae_CVPR_2022, cho2022unsupervised, Georgescu_2021_CVPR, Georgescu-TPAMI-2021, liu2018ano_pred, Markovitz_2020_CVPR, Morais_2019_CVPR, rodrigues2020multi, Sultani_2018_CVPR, wang2022jigsaw-vad, Zaheer_2022_CVPR}, anomaly detection in videos remains a difficult task. Utilizing conventional discriminative methods is challenging as the amount of video data captured exceeds our ability to label it manually. Furthermore, as abnormal events are far more infrequent than normal ones, the goal is to detect anomalies using only normal events.

We desire an unsupervised algorithm that can detect abnormal actions based on the context of past normal activities.
Moreover, we require the method to disregard nuisance parameters in the videos, such as background clutter, illumination changes, and other factors unrelated to the extent of the human activity normality.

This is why we only use skeleton data. This representation distills anomaly detection to human activity. 
Human pose sequences are highly semantic, intuitive, and very low dimensional signals, incorporating information regarding the extent of a
video’s normality~\cite{Markovitz_2020_CVPR, Morais_2019_CVPR, rodrigues2020multi}.
They also provide interpretability for the abnormality in the scene, as each human's pose sequence is given a normality score. A side-effect of working only with skeleton representation is that it can reduce privacy violation and bias that might be associated with appearance-based methods~\cite{privacy0, buet2022towards, WhereAreWeWithHumanPoseEstimation, privacy1, privacy2, raji2020saving, steed2021image}.

Previous pose-based methods build on autoencoders to reconstruct or predict future poses~\cite{Morais_2019_CVPR, rodrigues2020multi}. They use reconstruction error as a proxy for the normality score. However, autoencoders tend to generalize strongly, i.e., anomalies can be reconstructed as well as normal samples~\cite{gong2019memorizing}.
Markovitz \etal~\cite{Markovitz_2020_CVPR} suggested using a Dirichlet process mixture to determine the normality score. However, these scoring algorithms are ineffective for anomaly detection as they highly depend on the autoencoder's generalization capabilities and performance~\cite{score1, score0, score2}. The use of memory mechanisms~\cite{memory0, memory1} or multi-modal data~\cite{Georgescu-TPAMI-2021, multi0, multi1} can suppress this generalization to some extent but at the cost of additional memory consumption and computation.

In contrast, our end-to-end approach is based on a normalizing flows architecture, which learns the data distribution and scores the samples according to their likelihood instead of relying on reconstruction error.
We leverage the highly semantic pose representation to extend the architecture of normalizing flows, adapting it explicitly for human pose data using spatio-temporal graph convolution blocks.
Kirichenko \etal~\cite{kirichenko2020normalizing} show that while normalizing flows perform poorly when trained on pixel data, they can detect OOD images when trained on high-level semantic representations. But, unlike pixels, pose data holds substantial semantic value. Thus, abstracting videos to pose data forces normalizing flows to concentrate on semantic features.
To our knowledge, we are the first to propose a pose-based normalizing flows approach for anomaly detection. 

Utilizing low-dimensional pose data directly as a feature eliminates the need for autoencoders, used as feature extractors. This results in an incredibly lightweight network that runs in real-time ($\sim$1K parameters, 0.95G FLOPS, and inference speed of 189 FPS\footnote{On a single Nvidia Titan XP GPU.}), demanding negligible additional resources for any device capable of running pose estimation.
Note that the pre-trained pose estimator and tracker we use for preprocessing (we do not train or fine-tune this model) is of size 83M parameters.
But, as we are not limited to a specific pose estimator, we may benefit from improvements in pose estimation algorithms such as better accuracy and run-time optimization.

Our framework is as follows. Given a sequence of video frames, we use pose estimation to extract the keypoints of every person in each frame and use a pose tracker to track the skeletons across the frames. Eventually, each person in a clip is represented as a temporal pose graph.
Our network maps the training samples into a Gaussian-distributed latent space and calculates the probability of a human pose sequence.
Figure~\ref{fig:arch} provides an overview of this framework.

Our model works both in the usual unsupervised setting, where only normal data is given for training, and in the supervised setting, where both normal and abnormal training data are provided. In the supervised setting, we use our suggested normalizing flows model with a Gaussian Mixture Model prior, which forces the network to assign low probabilities to known abnormal samples.

We demonstrate our algorithm in the two settings. The first is the widely used ShanghaiTech Campus dataset~\cite{liu2018ano_pred}. In this unsupervised setting, the training data consists of only normal videos, and the test data consists of both normal and abnormal videos.
For the supervised setting, we use the recent synthetic UBnormal dataset~\cite{Acsintoae_CVPR_2022}, which consists of both normal and abnormal training data. We also test this dataset in the common unsupervised setting. 

Extensive experiments show that our model outperforms previous pose-based and appearance-based state-of-the-art methods for both settings. In addition, the ablation study shows our method is robust to noise and can generalize across different datasets. To summarize, we propose three key contributions:
\begin{itemize}
\item Introducing a normalizing flows architecture for both supervised and unsupervised video anomaly detection;
\item Extending normalizing flows networks to tackle the unique aspects of human pose data;
\item State-of-the-art AUC of 85.9\% for ShanghaiTech and 79.2\% for UBnormal anomaly detection benchmarks.
\end{itemize}
\section{Related Work}
\subsection{Video anomaly detection}
In recent years, numerous works tackled the problem of video anomaly detection using deep learning-based models. They deliver better performance by leveraging the powerful representation capabilities of deep neural networks. 

Specifically, structured representations, extracted using a pre-trained object detector~\cite{Georgescu_2021_CVPR, Georgescu-TPAMI-2021, wang2022jigsaw-vad} or pose estimator~\cite{Markovitz_2020_CVPR, Morais_2019_CVPR, rodrigues2020multi}, have attracted increased attention for their potential to get closer to the semantic concepts present in the anomalies. 
They use self-supervision on the extracted representations to learn semantic features by proxy tasks~\cite{cho2022unsupervised, Georgescu_2021_CVPR, Markovitz_2020_CVPR, rodrigues2020multi}, such as frame reconstruction and prediction, or by using out-of-domain data samples as anomalies, sometimes referred to as pseudo-anomalies~\cite{Georgescu-TPAMI-2021}. 

Georgescu \etal~\cite{Georgescu-TPAMI-2021} proposed a framework using object detection and optic flow as input data. They trained a set of autoencoders and classifiers using pseudo-abnormal examples to discriminate between normal and abnormal latent features.
Ristea \etal~\cite{Ristea_2022_CVPR} later suggested using a predictive convolutional attentive block to enhance the performance of the autoencoders. This block uses dilated convolutions with a masked center area and a reconstruction-based loss for the masked area.
Georgescu \etal~\cite{Georgescu_2021_CVPR} used self-supervised multi-task learning on object detection results using a single 3D convolutional backbone with multiple heads. However, these tasks are easy to solve, preventing the network from learning strong discriminative representations~\cite{wang2022jigsaw-vad}.
Thus, Wang \etal~\cite{wang2022jigsaw-vad} used an object-centric cube extraction and a spatial or temporal shuffling as a self-supervised task. They used a CNN-based network to recover the original sequence from its permuted version.

As these methods rely on RGB data, ethical concerns arise regarding anonymization and proven bias against groups of people based on their appearance. 
Steed \etal~\cite{steed2021image} showed that state-of-the-art unsupervised models trained on ImageNet automatically learn racial, gender, and intersectional biases. 
Buet \etal~\cite{buet2022towards} demonstrated on several datasets that there are discrepancies, sometimes crucial, between the reconstruction error of one group and another. This is especially troubling for anomaly detectors that use frame reconstruction as the normality score.
Furthermore, Liu \etal~\cite{privacy1} noted that models could be reverse-engineered, retrieving sensitive training data, including recognizable face images. Human-pose-based networks eliminate the possibilities of appearance-based discrimination or sensitive training data retrieval.

Nevertheless, video anomaly detection using human-pose data has not been overly explored. 
Morais \etal~\cite{Morais_2019_CVPR} proposed an anomaly detection method using an RNN to analyze pose sequences. They scored the videos according to the prediction error. 
Rodrigues \etal~\cite{rodrigues2020multi} later suggested a multi-timescale model capturing the temporal dynamics by making future and past predictions at different timescales.
Markovitz \etal~\cite{Markovitz_2020_CVPR} used a deep clustering network and a Dirichlet process mixture to determine the anomaly score. 
These methods rely on autoencoders, which although trained using only normal samples, have inherent generalization abilities that still make anomalous samples well reconstructed or predicted, especially for static data~\cite{memory0,memory1}.
In contrast, our method is based on a normalizing flows architecture, which can learn the density estimation of high-dimensional data. Therefore, our scoring function is the estimated probability of the input sample.

\subsection{Graph Convolutional Networks}
The graph relations of human poses are described using weighted adjacency matrices to represent the human skeleton as a temporal graph.
Yan \etal~\cite{yan2018spatial} and Yu \etal~\cite{yu2017spatio} proposed temporal extensions to graph convolution methods, with the former offering the use of separable spatial and temporal graph convolutions (ST-GCN), applied sequentially. We follow this ST-GCN block design.

\subsection{Normalizing Flows}
Normalizing flows (NF) are generative models that normalize complex real-world data distributions to “standard” distributions, usually Gaussian, using a sequence of invertible and differentiable transformations. The mathematical formulation for this architecture is introduced in section~\ref{Texts/background}.

Using normalizing flows for video anomaly detection has almost not been explored. Cho \etal~\cite{cho2022unsupervised} extracted video features using a reconstruction-based autoencoder (on RGB data). They then used normalizing flows networks to estimate the likelihood of the extracted features. This method greatly depends on the autoencoder’s learned representations, thus limiting its effectiveness.

In the image domain, Dinh \etal~\cite{dinh2016density} proposed RealNVP, a widely used generation network with an affine coupling layer.
Kingma \etal~\cite{kingma2018glow} later proposed Glow for further improvements using activation normalization and invertible 1×1 convolution. 
Izmailov \etal~\cite{izmailov2019semi} used normalizing flows for semi-supervised learning by using a Gaussian Mixture Model (GMM) as a prior and assigning each label to a different Gaussian mean and variance. The classification is done by calculating the probability of the sample conditioned on each label.

In the human pose domain, normalizing flows have only been suggested for generating synthesized pose data. Henter \etal~\cite{henter2020moglow} suggested using LSTMs in normalizing flows networks to synthesize 3D human motion data. Yin \etal~\cite{yin2021graph} extended this concept and used spatial GCNs and LSTMs to preserve temporal information.

\section{Method}
The overall structure of our algorithm is as follows.
Given a video sequence, we use a standard human pose detector and tracker to extract poses.
The poses are represented as space-time graphs and, using normalizing flows, 
are embedded into a Gaussian latent space. 
Finally, using Equation~\ref{log_p}, we can evaluate the probability of actions based on the training data. 

In what follows, we start with a brief overview of normalizing flows. Next, we introduce STG-NF, our adaptation of normalizing flow to work on space-time graphs. 
Finally, we show STG-NF in the supervised setting (i.e., when labeled abnormal examples are available during training).

\subsection{Normalizing Flows}
Normalizing flows~\cite{dinh2014nice} is an unsupervised model for density estimation defined as an invertible mapping $f\nolinebreak:\nolinebreak\mathcal{X}\nolinebreak\leftrightarrow\mathcal{Z}$ from the data space \(\mathcal{X}\) to the latent space \(\mathcal{Z}\). 
In most flow-based models, \(\mathcal{Z}\) has a tractable density, such as a spherical multivariate Gaussian distribution: \(\mathcal{Z} \sim \mathcal{N}(\mu, \Sigma)\). 

The function $f$ is implemented using a neural network parameterized by parameters $\theta$:
\[ z = f_\theta(x) = g_\theta^{-1}(x) \quad s.t.\quad x = g_\theta(z)\]  

We focus on functions where $f$ is composed of a sequence of transformations:
\[f=f_1\circ f_2\circ ...\circ f_K\]
Such a sequence of invertible transformations is also called a (normalizing) flow. They are parameterized by a neural network with architecture designed to ensure invertibility and efficient computation of the Jacobian determinants. 

The probability density (PDF) of the transformed variable is given by the change of variables formula:
\begin{equation}\label{eq_origin}
p_X(x) = p_Z(f(x))\cdot|det(\dfrac{df}{dx})|
\end{equation}
The Jacobian determinant captures the volume change between $X$ and $Z$. Applying logarithm and the chain rule:
\begin{equation}\label{log_p}
log \, p_X(x) = log \, p_Z(f(x)) + \sum_{i=1}^{K}log \,|det(\dfrac{df_i}{df_{i-1}})|
\end{equation}

The basic idea is to choose transformations $f_i$ whose Jacobian 
is a triangular matrix. For those transformations, the log-determinant is simple to calculate, enabling direct estimation of $p(x)$.
In comparison, other generative methods consider the probability distribution of the data, $p(x)$, to be intractable. As a result, sub-optimal scoring functions like reconstruction or prediction are used.
\label{Texts/background}

\subsection{Spatio-Temporal Graph Normalizing Flows}\label{STG-NF}


\begin{figure}
\centering
\includegraphics[width=0.4\textwidth]{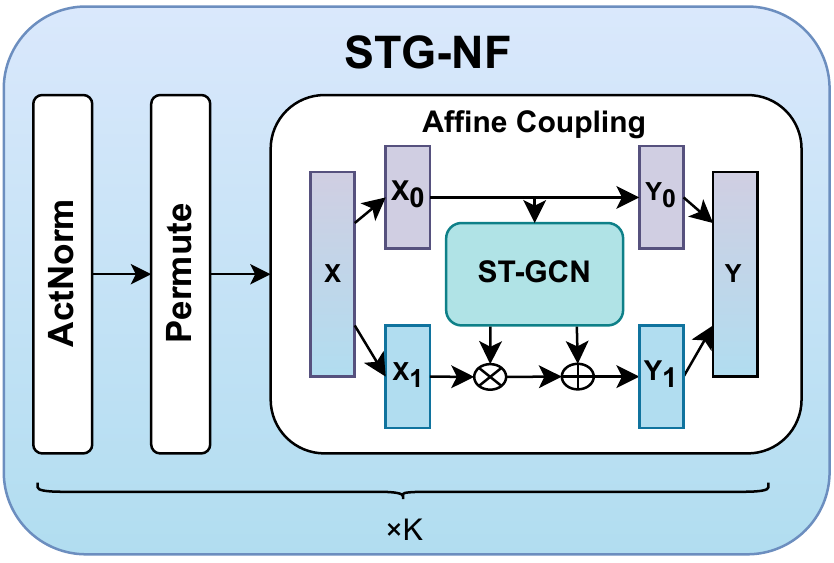}
\caption{\textbf{Spatio-Temporal Graph Normalizing Flows:} Our STG-NF module consists of $K$ flow steps, each with Actnorm, permutation, and spatio-temporal affine coupling layers. To leverage the semantics of the pose data, we use spatio-temporal graph convolutions in the affine layers. In these layers, we split the input
channel-wise and use one half to transform the other.
This architecture ensures easy calculation of the input data probability, $p(x)$.}

\label{fig:affine}
\end{figure}

Our normalizing flows-based model works directly on the pose sequence data. It learns an invertible mapping between the data distribution $P_X$ and a latent distribution $P_Z$, which is a Gaussian \(\mathcal{Z} \sim \mathcal{N}(\mu_{normal}, I)\), as demonstrated in Figure \ref{fig:arch}.
We leverage the compact pose representation and extend the image-oriented Glow~\cite{kingma2018glow} architecture to tackle pose sequences by using temporal graph convolutions in the affine layers. The architecture includes $K$ flow steps, each consisting of three layers: Actnorm, permutation, and an affine coupling layer. This architecture ensures invertibility, as each component in the network is bijective.

Actnorm~\cite{kingma2018glow} is an activation normalization layer that applies a scale and bias using data-dependent initialization, similar to batch normalization.
The permutation layer permutes the ordering of the input channels. We follow Glow~\cite{kingma2018glow} and use a learned 1×1 convolution.
This layer implements a reversible soft permutation by a linear transformation, which is initialized as a random rotation matrix.
The affine coupling layer splits the input $X\in\mathbb{R}^{BCTN}$ into two parts channel-wise, 
where B is the batch size, $C$ is the number of channels, $T$ is the segment length, and $N$ is the number of nodes. One half of the input is kept unchanged, while the other is affine transformed based on the other half. This leads to an easy reverse transformation and Jacobian calculation. 
To benefit from the rich semantics of the pose data, we use spatio-temporal graph convolutions (ST-GCN~\cite{yan2018spatial}). See Figure~\ref{fig:affine}.

The pose keypoints extracted from a video sequence (a time series of human joint locations) represent an undirected graph $G=(V, E)$. Each node $V=\{V_{it} \mid t\in[1, \tau],i\in[1, N]\}$ corresponds to a skeleton sequence with $N$ joints and $\tau$ frames, and each edge $E$ represents some relation between two nodes. 
There are many skeletal relations, such as anatomical (\eg the right shoulder and elbow are connected) and motion (\eg the left and right knees tend to move in opposite directions while walking). Spatial connections are defined by anatomical relations, and temporal connections are defined by connecting the same joints in successive frames. 
Using graph convolution, the output value for a single channel can be written as:

\begin{equation}
f_{out}(v_{it}) = \sum_{v_{lm} \in N(v_{it})} f_{in}(v_{lm}) \cdot w(v_{it}, v_{lm})
\end{equation}

where $N(v_{it})$ defines the connected nodes of $v_{it}$ and $w(v_{it}, v_{lm})$ the connection weight between nodes, as defined in the spatial and temporal adjacency matrices.

We train the model by minimizing the negative log-likelihood of the training samples:
\begin{equation}\label{nll}
L_{nll} = - log \, p_Z(f(x)) - log \, |det(\dfrac{df}{dx})|
\end{equation}
And assuming a Gaussian distribution \(\mathcal{N}(\mu_{normal}, I)\) in Equation~\ref{log_p} results in the following expression:
\begin{equation} \label{abnorm_eq}
\begin{aligned}
log \, p_Z(z) = Const &-\frac{1}{2}(z-\mu_{normal})^2  
\end{aligned}
\end{equation}
that is used to calculate $L_{nll}$ in Equation \ref{nll}.

\subsection{Supervised Anomaly Detection}

\begin{figure}
\centering
\includegraphics[width=0.48\textwidth]{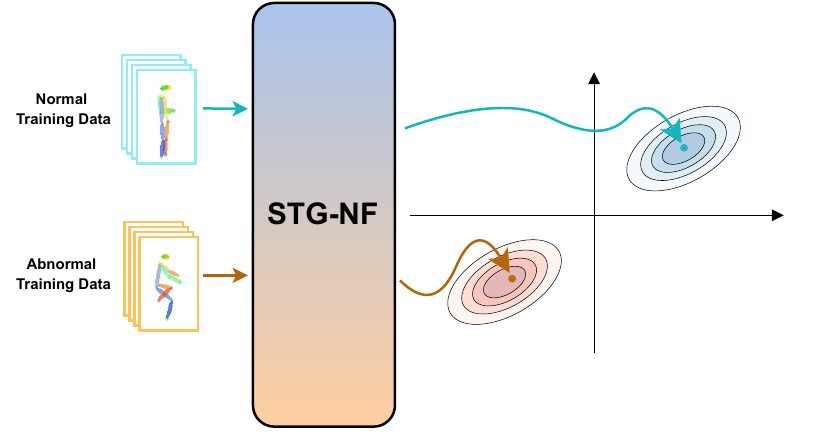}
\caption{\textbf{Supervised Anomaly Detection Overview:} We use a GMM prior for $P_Z$. We train by mapping pose sequences $X_i$ to latent representations with density \(\mathcal{N}(\mu_i, I)\), where $i$ depends on the label (normal or abnormal). During inference, we evaluate using $\mu_{normal}$. This prior forces the network to assign low probabilities to abnormal training samples.}
\label{fig:supervised}
\end{figure}

We extend our unsupervised normalizing flows anomaly detection method to a supervised setting. We use a Gaussian Mixture Model (GMM) with two components corresponding to the two classes (normal and abnormal) as a prior for our model. Figure~\ref{fig:supervised} demonstrates this extension.

During training, we map the pose sequence $X_i$ to a latent representation with density \(\mathcal{Z} \sim \mathcal{N}(\mu_i, I)\), where $i$ depends on the label of the pose sequence (normal or abnormal). Training is done the same as in the unsupervised setting, minimizing the negative log-likelihood using the new prior.

To force the network to assign low probabilities to abnormal samples, we choose $\mu_i$ such that:
\[|\mu_{normal}-\mu_{abnormal}| \gg 0 \]

During inference, we calculate the normality probability score using $\mu_{normal}$, the same as in the unsupervised setting (Equation~\ref{nll}). Thus, for \(f(x_{normal})\sim\mathcal{N}(\mu_{normal}, I)\) we get high probabilities and for \(f(x_{abnormal})\sim\mathcal{N}(\mu_{abnormal}, I)\) we get low probabilities.
\section{Experiments}
\begin{figure}
    \centering
    \begin{subfigure}[b]{0.15\textwidth}
        \includegraphics[width=\textwidth]{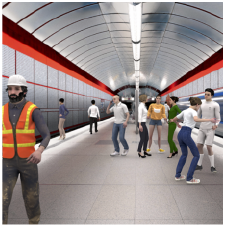}
        \caption{}
        \label{fig:UBnormal_a}
    \end{subfigure}
    \hfill
    \begin{subfigure}[b]{0.15\textwidth}
        \includegraphics[width=\textwidth]{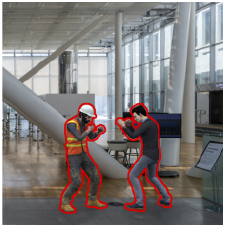}
        \caption{} 
        \label{fig:UBnormal_b}
    \end{subfigure}
    \hfill
    \begin{subfigure}[b]{0.15\textwidth}
        \includegraphics[width=\textwidth]{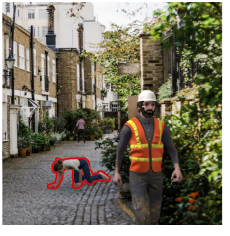}
        \caption{} 
        \label{fig:UBnormal_c}
    \end{subfigure}
\caption{\textbf{UBnormal Examples:} Examples from various scenes included in the synthetic UBnormal dataset~\cite{Acsintoae_CVPR_2022}. Figure (a) demonstrates a normal scene. Figures (b) and (c) show abnormal actions, emphasized through red contours.} 
\label{UBnormal_examp}
\end{figure}

We evaluate our model in two different settings, using two datasets. The first setting is the common unsupervised anomaly detection setting, where the training data consists of only normal videos, and the test data consists of both normal and abnormal videos. The second supervised setting consists of both normal and abnormal training data.
We evaluate the ShanghaiTech Campus dataset~\cite{liu2018ano_pred} on the unsupervised setting, and the recent synthetic UBnormal dataset~\cite{Acsintoae_CVPR_2022} on both the unsupervised and supervised settings.
Figure~\ref{UBnormal_examp} shows examples from UBnormal dataset.

\begin{figure}
\centering
\includegraphics{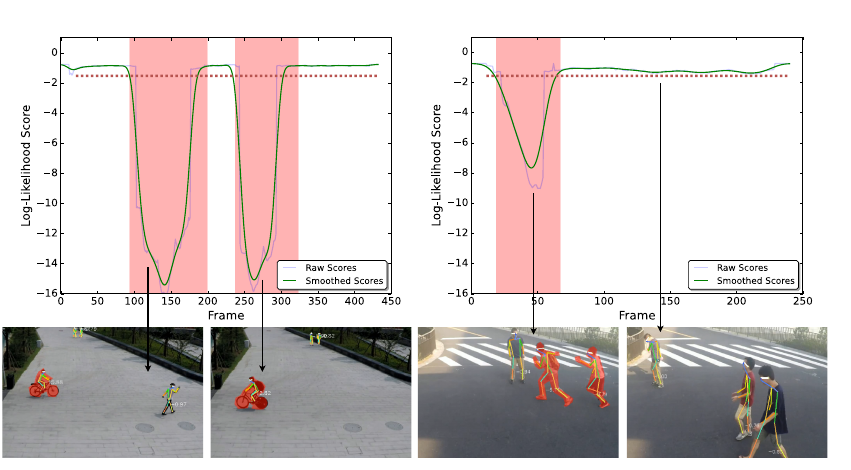}
\caption{\textbf{Log-likelihood Score Examples:} Score examples for ShanghaiTech dataset videos - ground truth anomalous frames are marked red, dashed red line simulates the threshold. Each frame is scored according to the minimal pose score in that frame. Our method correctly flags the anomalies in both space and time.}
\label{fig:demo}
\vspace{-3px}
\end{figure}

For evaluation, a score is calculated for each frame individually. The dataset score is the area under the ROC curve for concatenating all frame scores in the test set, which is the most common metric, sometimes called Micro AUC. An example from ShanghaiTech dataset is shown in Figure~\ref{fig:demo}.
To assess the localization of anomalies, we evaluate our model using a region-based detection criterion (RBDC) and a track-based detection criterion (TBDC), as proposed by Ramachandra~\etal \cite{ramachandra2020street}. 
RBDC evaluates each detected region by considering its Intersection-over-Union (IOU) with the corresponding ground-truth region, marking it as a true positive if the overlap exceeds a threshold $\alpha$.
TBDC assesses the accuracy of tracking abnormal regions over time, considering a detected track as a true positive if the number of detections within the track surpasses a threshold $\beta$. Following the conventions established in~\cite{ssmtl++, Georgescu-TPAMI-2021, ramachandra2020street,  Ristea_2022_CVPR}, we set $\alpha$ and $\beta$ to $0.1$.
\begin{table*}
\centering
\begin{tabular}{lcccc}

\toprule

\textbf{Model}  & \centering\textbf{AUC} &\centering\textbf{ShanghaiTech-HR AUC} &\textbf{RBDC} &\textbf{TBDC}\\
\midrule

Morais \textit{et al.}$^{\ast}$~\cite{Morais_2019_CVPR}
&73.4 & 75.4 & - & -\\

Rodrigues \textit{et al.}$^{\ast}$~\cite{rodrigues2020multi}
&76.0 &  77.0 & - & -\\

Markovitz \textit{et al.}$^{\ast}$~\cite{Markovitz_2020_CVPR}
&76.1 &  - & - & -\\

\midrule


Zaheer \textit{et al.}~\cite{Zaheer_2022_CVPR} &79.6 & - & - & -\\


Georgescu \textit{et al.}$^{\dag}$~\cite{Georgescu-TPAMI-2021}+SSPCAB~\cite{Ristea_2022_CVPR} &83.6 & - & 40.6 & 83.5\\

Barbalau \textit{et al.}$^{\dag}$~\cite{ssmtl++} (SSMTL++v2) &83.8 & - & 47.1 & \textbf{85.6}\\

Wang \textit{et al.}$^{\dag}$~\cite{wang2022jigsaw-vad}   &84.2 & 84.7 & 22.4 & 60.8\\
\midrule 
\textbf{Ours}
& \textbf{85.9} & \textbf{87.4} & \textbf{52.1} & 82.4\\

\bottomrule
\end{tabular}

\caption{
\textbf{ShanghaiTech Results:} Frame-level AUC, RBDC, and TBDC comparison on ShanghaiTech. 
ShanghaiTech-HR is a subset that only contains human-related abnormal events. 
[$^{\ast}$] also uses pose data as input,
[$^{\dag}$] uses pre-trained object detection results as input.
}
\label{Tab:shanghai}
\end{table*} 
\subsection{Implementation Details}
\textbf{Data Preprocessing}. To detect skeletons in each video frame independently, we utilize AlphaPose~\cite{li2019crowdpose} with YOLOX~\cite{yolox2021} detector. We then use PoseFlow ~\cite{xiu2018poseflow} to track the skeletons across a video. After that, we  divide each pose sequence into fixed-length segments using a sliding-window approach. 
We used a segment window $\tau=24$ frames for ShanghaiTech and $\tau=16$ frames for UBnormal, as the pose estimation and tracking results of UBnormal are less accurate than ShaghaiTech's. 
Finally, we normalize each pose segment to have zero mean and unit variance.
During inference, we score each pose segment individually. For frames with more than one person, we take the minimal score over all the people in the frame:
\[ Score_t = \min_{\forall x \in X_t} log \,  p_X(x)\]

\textbf{Training}. 
For the unsupervised setting, we use the prior \(\mathcal{N}(3, I)\), and for the supervised setting we use \(\mathcal{N}(10, I)\)  for normal samples and  \(\mathcal{N}(-10, I)\) for abnormal samples.
To keep our network slim, we use $K=8$ flow steps and a single standard block of ST-GCN per flow step. Experiments showed that using more blocks resulted in minor performance gain compared to the size cost.
The supplementary includes additional implementation details and ablation studies (\eg performance using different flow steps $K$ and segment window size $\tau$).
\subsection{Unsupervised Anomaly Detection}
\textbf{ShanghaiTech}. The ShanghaiTech dataset is one of the largest datasets for video anomaly detection, containing videos from 13 cameras around ShanghaiTech University campus. It consists of 330 training videos with only normal events and 107 test videos with both normal and abnormal events, annotated at frame and pixel levels.
A few examples of human anomalies in the dataset are running, fighting, and riding bikes. The videos contain various people in each scene, with challenging lighting and camera angles.

We evaluate our method on ShanghaiTech and ShanghaiTech-HR, which is a subset of ShanghaiTech that only contains human-related abnormal events~\cite{Morais_2019_CVPR,rodrigues2020multi} (there are a total of six non-human anomalies test videos that are neglected from the original ShanghaiTech dataset).
We report results in Table~\ref{Tab:shanghai}. As can be seen, our method significantly outperforms other pose-based methods. Moreover, our method ranks first among appearance-based methods, even in the standard ShanghaiTech setting which includes some non-human anomalies.
Furthermore, we achieve SOTA results for RBDC and on-par results for TBDC, which indicates we can accurately localize anomalies.
Thus, it's possible to benefit the numerous advantages of using only pose data, without sacrificing performance.
\subsection{Supervised Anomaly Detection}
\begin{table}
\centering

\begin{tabular}{lcccc}

\toprule

\textbf{} & \textbf{Model}  &\textbf{AUC} &\textbf{RBDC} &\textbf{TBDC}\\
\midrule
 
\multirow{6}{*}{\begin{sideways}Unsupervised\end{sideways}} 
&Markovitz \textit{et al.}$^{\ast}$~\cite{Markovitz_2020_CVPR} &52.0 & - & -\\
&Georgescu \textit{et al.}$^{\dag}$~\cite{Georgescu-TPAMI-2021} &59.3 & 21.9 & 53.4\\
&Georgescu \textit{et al.}$^{\dag}$~\cite{ssmtl++} &62.1 & 25.6 & \textbf{63.5}\\
&Wang \textit{et al.}$^{\dag}$~\cite{wang2022jigsaw-vad}  &56.4 & 11.8 & 36.8\\
\cmidrule(lr){2-5}
&\textbf{Ours - Unsupervised} &\textbf{71.8} &\textbf{31.7} & 62.3\\
 
\midrule
\multirow{8}{*}{\begin{sideways}Supervised\end{sideways}} 
 & Georgescu \textit{et al.}$^{\dag}$~\cite{Georgescu-TPAMI-2021} &61.3 & 25.4 & 56.3\\
 & Bertasius \textit{et al.}$^{\dag}$~\cite{bertasius2021space} &68.5 & 0.04 & 0.05\\
 & Yan \textit{et al.}$^{\ast}$~\cite{yan2018spatial} & 78.1 & \textbf{46.2} & 69.9\\
 & Shi \textit{et al.}$^{\ast}$~\cite{2sagcn2019cvpr} & 74.1 & 42.9 & 73.4\\
 & Cheng \textit{et al.}$^{\ast}$~\cite{cheng2020shiftgcn} & 64.6 & 40.2 & 70.5\\
 & Liu \textit{et al.}$^{\ast}$~\cite{liu2020disentangling} & 77.8 & 43.9 & 70.8\\
 \cmidrule(lr){2-5}
 & \textbf{Ours - Supervised} &\textbf{79.2} & 43.6 &\textbf{75.6}\\


\bottomrule
\end{tabular}

\caption{
\textbf{UBnormal Results:} Frame-level AUC, RBDC, and TBDC comparison on UBnormal supervised and unsupervised settings. 
[$^{\ast}$] also uses pose data as input,
[$^{\dag}$] uses pre-trained object detection results as input.
}
\label{Tab:ubnormal}
\end{table} 
\textbf{UBnormal}. The UBnormal dataset is a new and synthetic supervised open-set benchmark containing both normal and abnormal actions in the training set. It contains 268 training videos, 64 validation videos, and 211 test videos and is also annotated at both frame and pixel levels. 

Some scenes in the dataset include foggy and night scenes. The pose detector overcame these difficult conditions and accurately estimated the poses in such scenes, showcasing the benefits of a non-appearance-based model, which disregards nuisance parameters like illumination and background clutter. 
However, for scenes with fire and smoke as abnormal events, appearance-based models gain an edge. Detecting these events from pose data relies on people's reactions to such scenarios.

We evaluate UBnormal using both the unsupervised and the supervised settings.
Pixel-level annotations were used to extract the abnormal pose data for training.
We report results in Table~\ref{Tab:ubnormal}. Again, our model outperforms previous methods in both settings by a large margin, with the supervised approach performing better on this imbalanced dataset (mostly normal actions).
However, the supervised setting requires access to abnormal data, which might be challenging to obtain.
Regarding localization in the unsupervised setting, as in ShanghaiTech, we achieve SOTA results for RBDC and on-par results for TBDC. For the supervised setting we outperform image-based methods by a large margin. Compared to pose-based methods, we achieve on-par results for RBDC and SOTA for TBDC. These results provide strong evidence of our ability to accurately localize anomalies.

\subsection{Ablation Study}
We conduct a number of ablation studies on the ShanghaiTech and UBnormal datasets. First, we evaluate the contribution of the ST-GCN network in the affine layer on performance. Next, we evaluated the spatial and temporal contributions to the algorithm. In another experiment, we consider the generalization of our model by training it on one dataset and testing it on the other. Finally, we assess the robustness of our model to noisy data. Additional ablation studies are presented in the supplemental material (\eg modeling performance when using different pose estimators or using limited data for training).

\paragraph{Affine Layer:} We replace our suggested spatio-temporal convolutional affine layer with a fully-connected affine layer and evaluate it on ShanghaiTech dataset. As a result, the AUC decreased from $85.9\%$ to $77.0\%$. Similarly, using standard convolutions reduced our score to $78.8\%$, verifying the contribution of our graph-oriented architecture.

\paragraph{Space-Time:} We experiment with different spatial and temporal relations. To understand the temporal effect, we set the input segment length to $\tau=1$, which yielded an AUC of $75.1\%$ (further ablation of different segment lengths can be seen in the supplementary). We also reduced the spatial relations by using an adjacency matrix $A=I$, where each joint is only connected to itself (temporally), which resulted in an AUC of $83.9\%$.
Compare these results to the combined score of $85.9\%$. We conclude that spatial connections are less important than temporal ones. In fact, observing the temporal behavior of each joint over time provides most of the information for anomaly detection.
\begin{table}
\centering

\begin{tabular}{c c | ccc}

\toprule

&  & \multicolumn{3}{c}{\textbf{Train}}\\

&  & ShanghaiTech & UBnormal & Both\\

\midrule
 
\multirow{3}{*}{\centering\begin{sideways}\textbf{Test}\end{sideways}} & ShanghaiTech &85.9 &84.0 & \textbf{86.0}\\
\cmidrule(lr){2-5}

& UBnormal &69.0 &\textbf{71.8} & 71.6 \\

\bottomrule
\end{tabular}

\caption{\textbf{Cross-Dataset Generalization:} 
Frame-level AUC for models trained on one dataset and tested on the other or trained using both datasets. 
ShanghaiTech is real surveillance footage, and UBnormal is synthetic.
Using pose data mitigates this appearance dissimilarity.
There is only a slight performance degradation compared to using the same dataset for training and testing.
}
\label{Tab:comb_ablat}
\end{table} 

\paragraph{Cross-Dataset Generalization:} Despite the appearance differences between UBnormal (synthetic) and ShanghaiTech (real surveillance footage) datasets, the actions considered normal in both datasets are similar. Our model, relying on pose data that abstracts appearance, can be trained on one dataset and tested on the other.
As can be seen in Table ~\ref{Tab:comb_ablat}, when training on one dataset and testing on the other, there is only a slight degradation in performance of up to $3\%$ in the AUC score, compared to using the same dataset for both training and testing. This suggests that it should be possible to train a model using synthetic data only (which is sometimes easier to acquire and label) and evaluate it in real-life scenarios.
Furthermore, we see a slight improvement in using both training sets on ShanghaiTech, and a slight decrease on UBnormal. 
These findings validate our model's robust understanding of human actions across different data domains.

\begin{figure}
\centering
\includegraphics[width=0.47\textwidth]{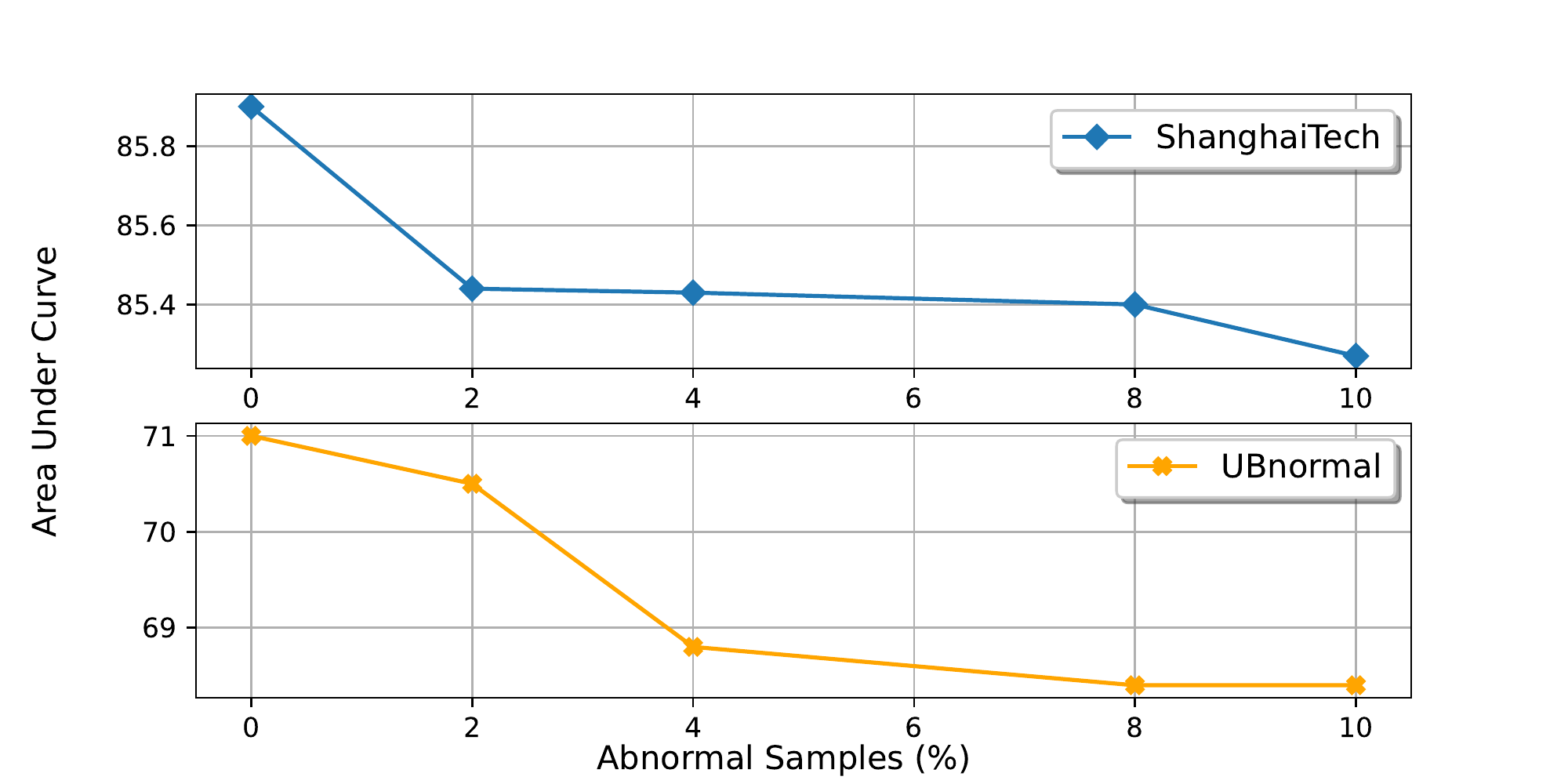}
\caption{\textbf{AUC for Training with Noisy Data:} Performance of unsupervised models trained on ShanghaiTech (top) or UBnormal (bottom) datasets, when a percentage of abnormal mislabeled samples are added from UBnormal train-set. Our model is robust to a significant amount of noise.}
\label{fig:noise}
\end{figure}

\paragraph{Noise Robustness:} In real-life scenarios, determining whether a scene contains only normal actions is difficult. Thus, robustness to noise is an essential property of anomaly detection models.
We work in the unsupervised setting, contaminating the normal training data with various amounts of abnormal examples from the UBnormal training set.
Figure~\ref{fig:noise} shows that our model loses less than $3\%$ AUC score when $5\%$ of the training data are abnormal examples, which demonstrates our model's robustness to noise.

\paragraph{Zero-Shot Anomaly Detection:} In the spirit of noise robustness, we propose a novel anomaly detection setting without relying on a clean training set of only normal samples. 
As our model learns the data distribution, we hypothesized it would assign higher probabilities to frequent (normal) actions than infrequent (abnormal) ones on the training data. 
We trained our network on the ShanghaiTech test set (without using the labels), where normal actions are more frequent. This setting might simulate online training, where we do not use annotations or filter the training data.
The evaluation was done normally on the test set, comparing each frame’s $p(x)$ to the ground truth, resulting in a high AUC score of $83.8\%$. This result indicates a promising ability to distinguish abnormal events without training on a clean dataset of only normal samples.

\subsection{Failure Cases}
\begin{figure}
    \centering
    \begin{subfigure}[b]{0.15\textwidth}
        \includegraphics[width=\textwidth]{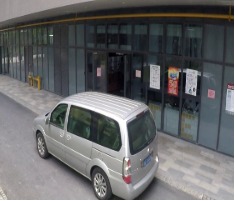}
        \caption{}
        \label{fig:fail_a}
    \end{subfigure}
    \hfill
    \begin{subfigure}[b]{0.15\textwidth}
        \includegraphics[width=\textwidth]{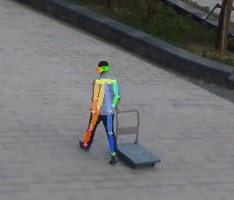}
        \caption{} 
        \label{fig:fail_b}
    \end{subfigure}
    \hfill
    \begin{subfigure}[b]{0.15\textwidth}
        \includegraphics[width=\textwidth]{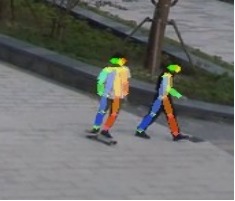}
        \caption{} 
        \label{fig:fail_c}
    \end{subfigure}
\caption{\textbf{Failure Cases:} Examples from ShanghaiTech dataset. (a) Our model does not handle non-person-related anomalies. (b) Using only pose data, dragging an anomalous object may seem like ordinary walking. (c) Due to the complex camera angles in the dataset, some actions, like skating, resemble normal walking.} \label{fail_examp}
\end{figure}

Figure~\ref{fail_examp} shows some failure cases, together with the extracted poses. Our model does not handle non-person-related anomalies, such as cars crossing a frame, demonstrated in \ref{fig:fail_a}. Moreover, as our method relies only on poses for detecting anomalies, a poorly extracted pose might be considered an anomaly. This limitation becomes more significant in the case of low-quality videos. It prevents us from evaluating our method on UCSD Ped1/Ped2~\cite{Ped}, popular datasets whose video resolution is too low to detect skeletons.
Furthermore, some anomalies in the ShanghaiTech dataset are people dragging items. Using only pose data, these actions seem like normal walking, as seen in \ref{fig:fail_b}.

In addition, due to camera angles, there is substantial variability in the pose's viewpoints. For instance, in the ShanghaiTech dataset, skating is considered anomalous. Because of some scenes' view angles, the movements of people walking (an action considered normal) toward the camera from afar are subtle, resulting in a pose sequence similar to a skater. This example is presented in \ref{fig:fail_c}.
\section{Ethical Considerations}
Anomaly detection algorithms for surveillance are used extensively because they help preserve safe environments. 
However, these algorithms invade privacy and might introduce bias based on appearance~\cite{buet2022towards, WhereAreWeWithHumanPoseEstimation, raji2020saving, steed2021image}. 
On top of that, deep-learning models can be reversed-engineered to reveal their training samples~\cite{privacy0, privacy1, privacy2}.

Our method uses human pose data only, thus enforcing fair decisions based only on the actions of the subject and not his appearance. In addition, using human pose data permits masking the identity of people in the training data. As our training data is human poses, we can completely mask the identity of the people in the dataset and guarantee privacy and anonymization.

We believe our work is a step in the right direction that will allow society to benefit from the advantages of surveillance systems while mitigating some of their shortcomings.

\section{Conclusions}
We propose a new human pose anomaly detection algorithm, that works on a space-time graph representation of human pose. The algorithm is based on a new Spatio-Temporal Graph Normalizing Flows (STG-NF) architecture, which is our adaption of normalizing flows to work with space-time graph data. 
By design, our model focuses on pose only and ignores nuisance parameters such as appearance or lighting. As a result, it is very lightweight, and prevents bias against distinct minority groups.
Extensive experiments using unsupervised and supervised settings show our method's robustness and generalization, which achieves state-of-the-art results on the ShanghaiTech and UBnormal benchmarks.
\section{Acknowledgements}
We thank Ofir Abramovich and Joni Alon for their help throughout this work. This work was partly funded by the Israel Innovation Authority grant number 75795, and Motorola Solutions.
{\small
\bibliographystyle{ieee_fullname}
\bibliography{egbib}
}
\clearpage

\section{Supplementary}
The following sections include additional information about our method. 

Section~\ref{ablation} presents further ablation studies conducted to evaluate our model. Section~\ref{implementation} provides implementation details of our method, and Section~\ref{baseline_imp} describes the implementation details of the baseline methods used. 
Section ~\ref{results} provides examples of our method's performance from the ShanghaiTech and UBnormal datasets.

\label{introduction}
\section{Ablation Study - Cont.}
\begin{figure}
\centering
\includegraphics[width=0.5\textwidth]{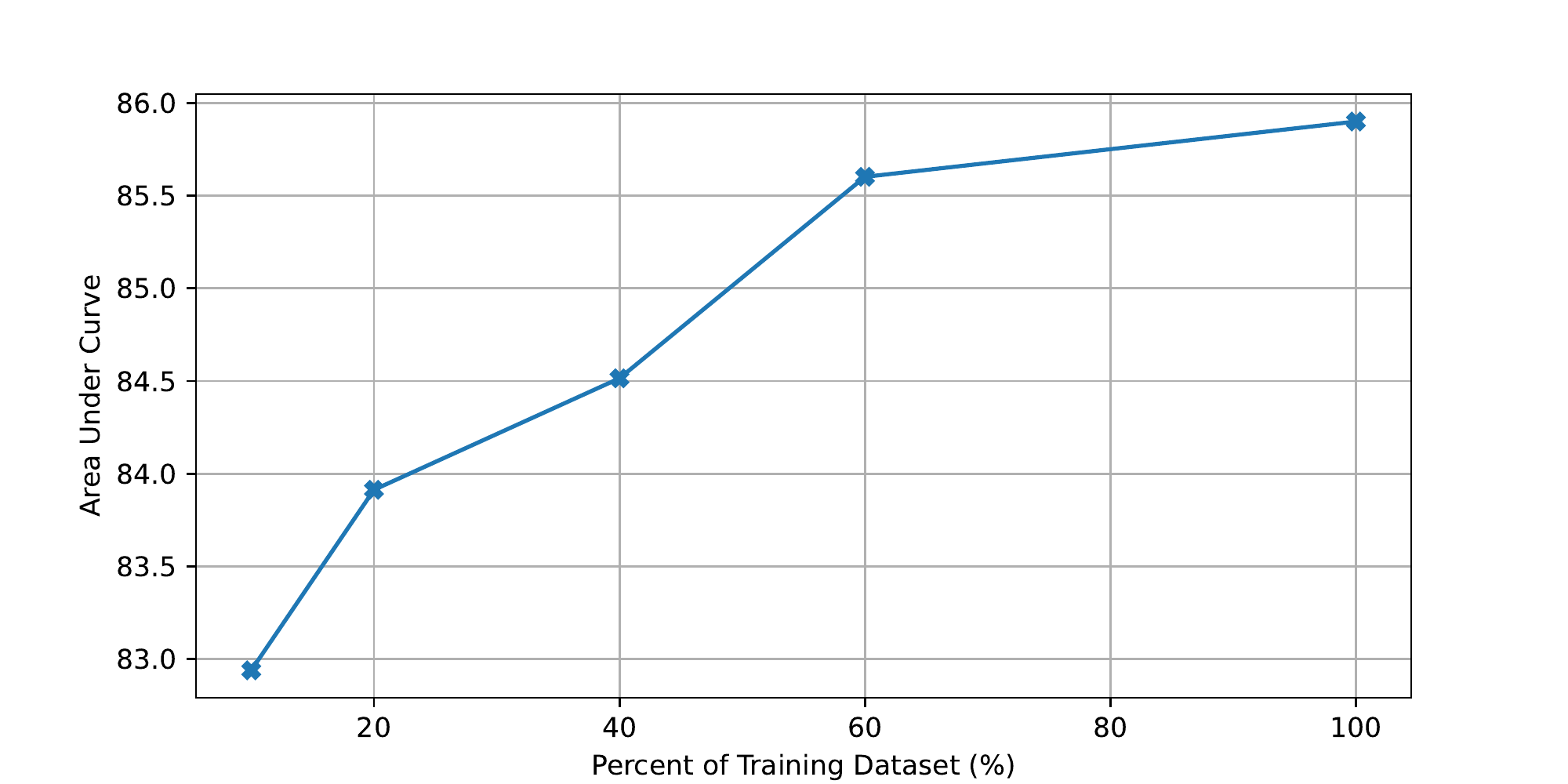}
\caption{\textbf{AUC Score for Partial Data Training:} Performance of unsupervised models trained on ShanghaiTech for different percentages of training data. Our model achieves near state-of-the-art performance using only a limited amount of training samples.}
\label{fig:partial}
\end{figure}

In this section, we provide further ablation experiments used to evaluate different model components:

\paragraph{Partial Data Training:} Acquiring normal training data might not be easy. Thus, evaluating the performance of our method using limited training data is essential. We use different sized subsets of the ShanghaiTech dataset $S_i$, where $S_i\subset S_{i+1}$ for $|S_i|<|S_{i+1}|$. 
Figure \ref{fig:partial} shows our method's performance using the different subsets.
Our model losses less than 3\% AUC using only 10\% of the training data. This is a side benefit of our compact model, which includes only $\sim1K$ parameters. In addition, as most actions considered normal in the dataset are walking, a small portion of the dataset is enough to assign high probability to this aciton. Thus, limited training data suffices for near state-of-the-art performance.

\begin{figure}
\centering
\includegraphics[width=0.5\textwidth]{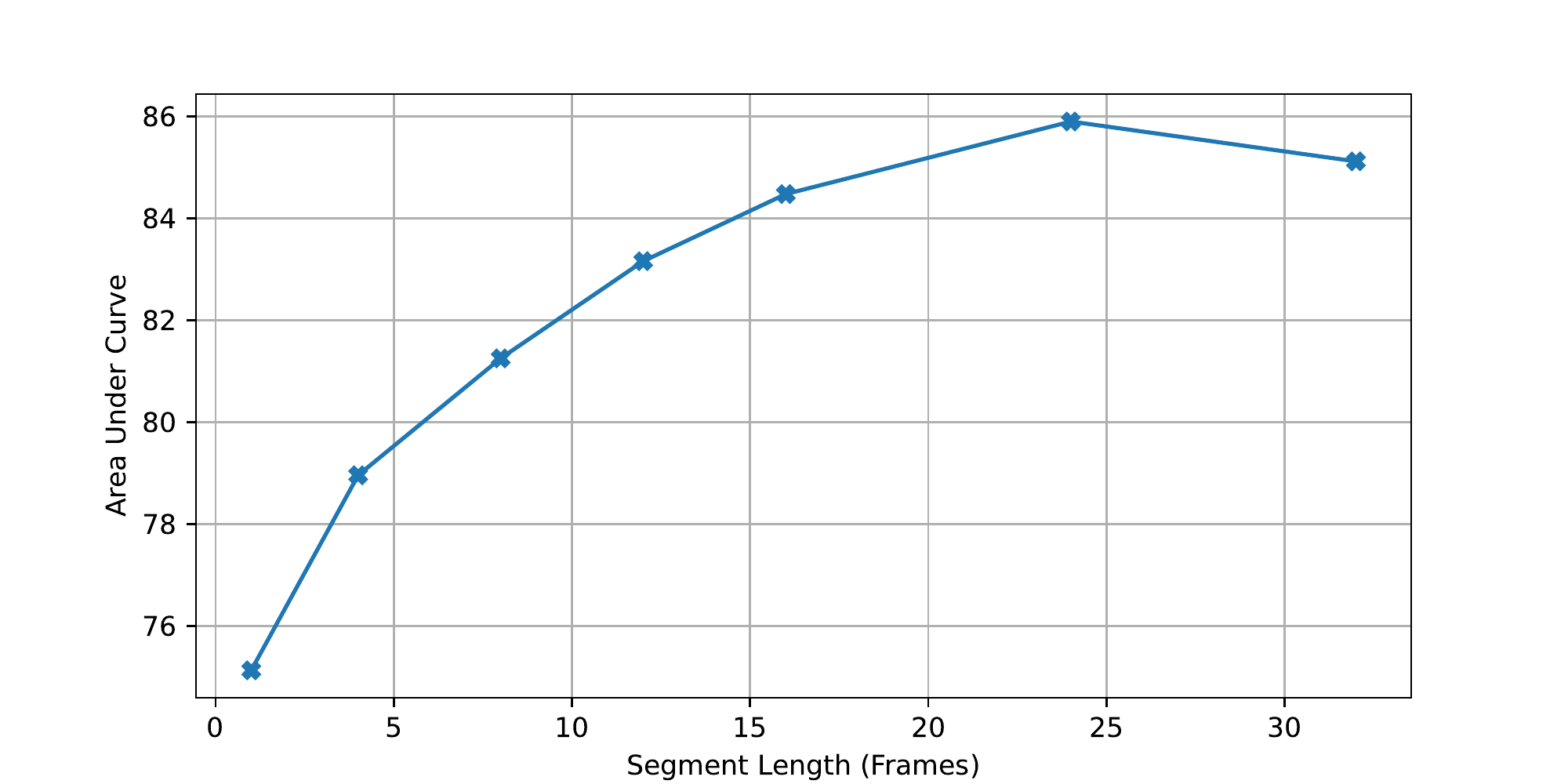}
\caption{\textbf{AUC Score for Different Segment Lengths:} Performance of unsupervised models trained on ShanghaiTech with different segment lengths. A larger segment window is preferable, subject to the pose estimation and tracking ability to output long human pose segments.}
\label{fig:segment}
\end{figure}

\paragraph{Segment Window:} We explore the effect of different pose segment windows $\tau$ on the performance of our model. As can be seen in Figure \ref{fig:segment}, using longer segment windows results in better performance, up until a time limit $\tau=T$. We believe this limit is affected by the performance of the human pose estimator and tracker. In general, a larger $\tau$ is desirable, so complex actions could be better learned. 
Thus, we conclude that using a larger segment window is preferable, subject to the pose estimation and tracking ability to output long human pose segments.

\begin{figure}
\centering
\includegraphics[width=0.5\textwidth]{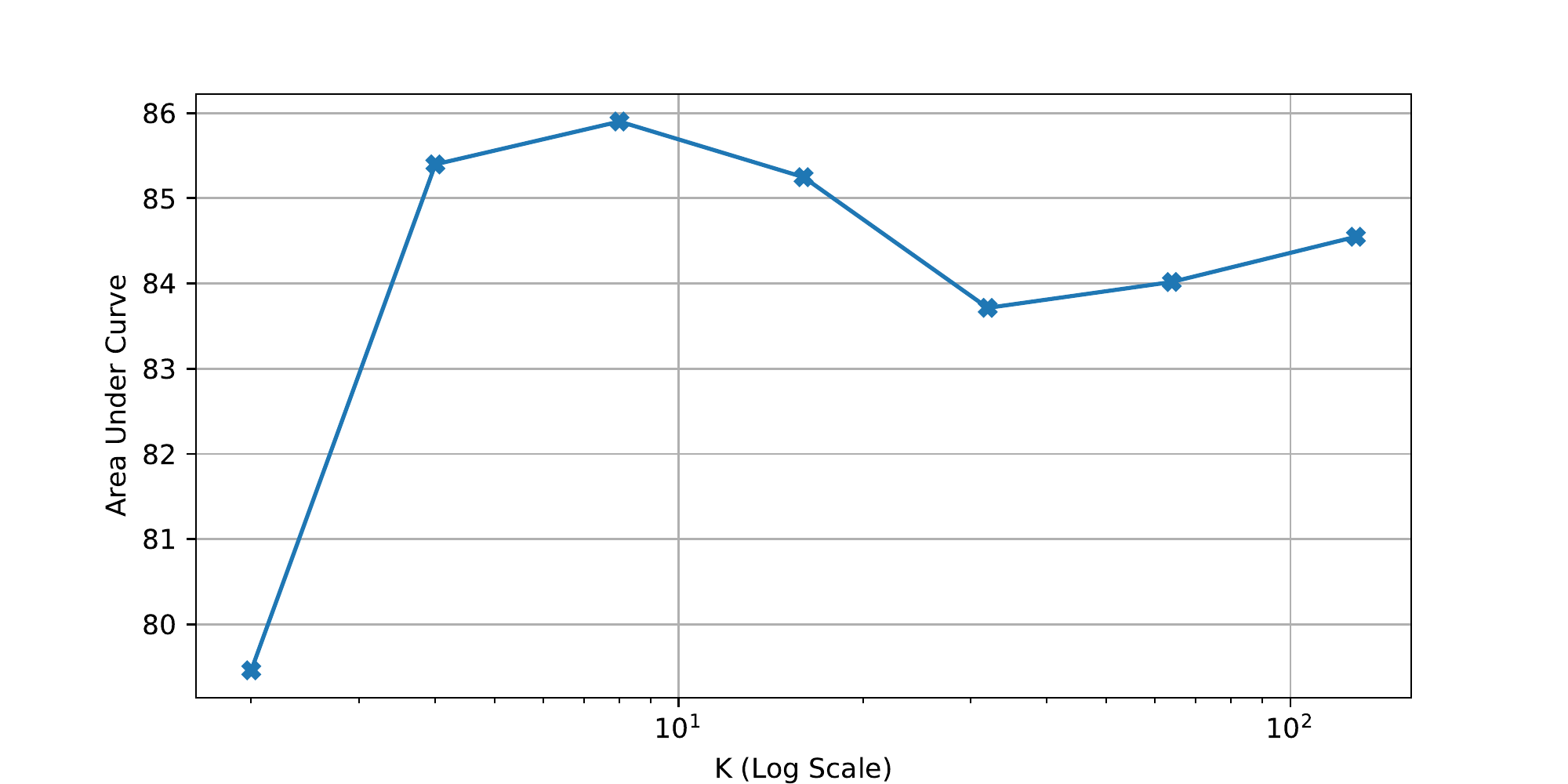}
\caption{\textbf{AUC Score for Different Numbers of Flows:} Performance of unsupervised models trained on ShanghaiTech with a different number of flows $K$. $K=8$ results in state-of-the-art performance. }
\label{fig:flows}
\end{figure}

\paragraph{Number of Flow Layers:} We explore our model's performance using different flow layers $K$ on the ShanghaiTech dataset. As demonstrated in Figure \ref{fig:flows}, using only $K=8$ results in state-of-the-art performance. We believe that for more complex datasets, a larger $K$ might be needed.

\begin{figure}[t]
\centering
\includegraphics[width=0.47\textwidth]{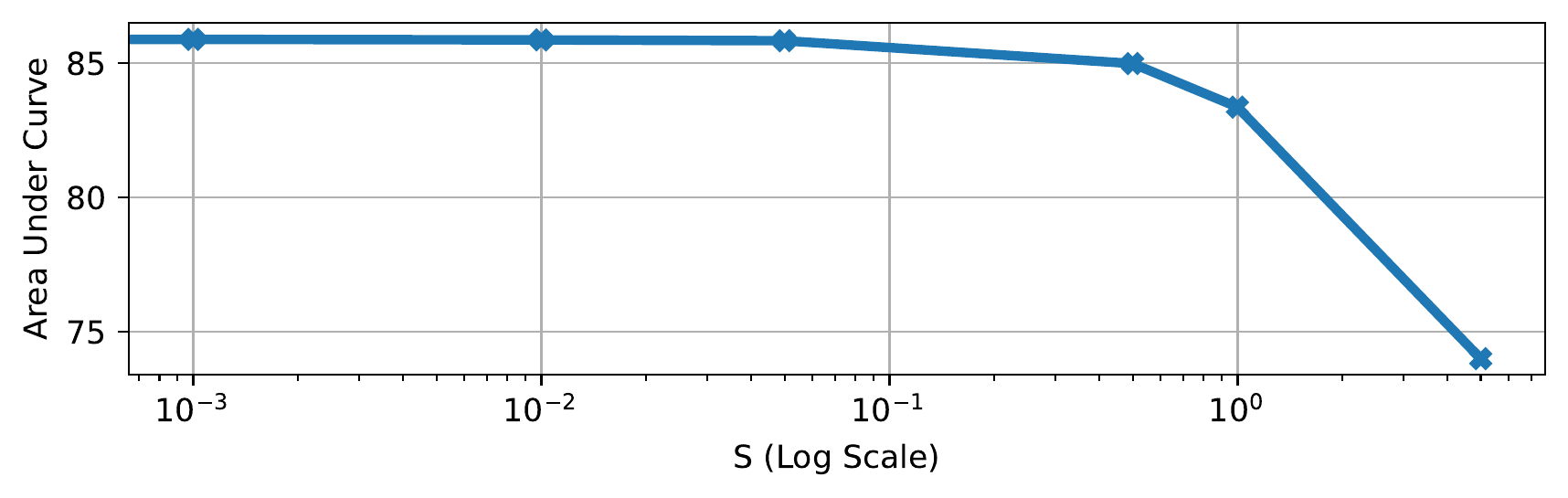}
\caption{\textbf{AUC for Noisy Keypoints:} Performance of models trained on ShanghaiTech, adding various scales of Gaussian noise.}
\label{fig:detect_noise}
\end{figure}

\paragraph{Pose Estimation Errors:} To analyze the impact of the pose estimators' errors, we modeled the estimators errors by adding to each keypoint Gaussian noise. After the pose normalization, we added varying scales $S$ of noise $S\cdot z$  where $z \sim \mathcal{N}(0, I)$. As shown in Figure~\ref{fig:detect_noise}, our model is robust to a significant amount of keypoints noise, which implies good performance when using less accurate pose estimators.

\label{ablation}
\section{Implementation Details}
Our method is implemented in PyTorch, and all experiments were conducted on a single NVIDIA Titan Xp GPU.

\textbf{Architecture}. We observed the best results when using the unit adjacency matrix in the affine layers, where each joint has an equal effect on the others. 
We use a segment window $\tau = 24$ frames for ShanghaiTech and $\tau = 16$ frames for UBnormal, as the results of the pose estimation and tracking of UBnormal are less accurate than ShaghaiTech.

\textbf{Training}. We optimize the normalizing flows network parameters with an Adam optimizer, learning rate $5*10^{-4}$, momentum 0.99, for eight epochs with batch size 256. 

For the unsupervised setting, we use the prior \(\mathcal{N}(3, I)\), and for the supervised setting we use
\(\mathcal{N}(10, I)\) 
for normal samples and 
\(\mathcal{N}(-10, I)\)
for abnormal samples, ensuring:
\[|\mu_{normal}-\mu_{abnormal}| \gg 0 \]
\label{implementation}
\section{Baseline Implementation Details}
The evaluation of the pose-based methods was conducted using their publicly available implementation\footnote{\url{https://github.com/amirmk89/gepc}}\footnote{\url{https://github.com/yysijie/st-gcn}}\footnote{\url{https://github.com/kchengiva/Shift-GCN}}\footnote{\url{https://github.com/lshiwjx/2s-AGCN}}\footnote{\url{https://github.com/kenziyuliu/ms-g3d}}. The training was done using the same pose data.

Similarly, the evaluation of the Jigsaw anomaly detection was conducted using their implementation\footnote{\url{https://github.com/gdwang08/Jigsaw-VAD}}.
The training was done using default parameters used by the authors, and changes were only made to adapt the data loading portion of the models to our datasets.

As we couldn't reproduce some of the results, thus we took the original AUC scores from the original paper. 
In addition, some results were taken from~\cite{ssmtl++}.\label{baseline_imp}
\section{Additional Results}
\begin{figure*}
\centering
    \begin{subfigure}{\textwidth}
    \includegraphics[width=1.0\textwidth]{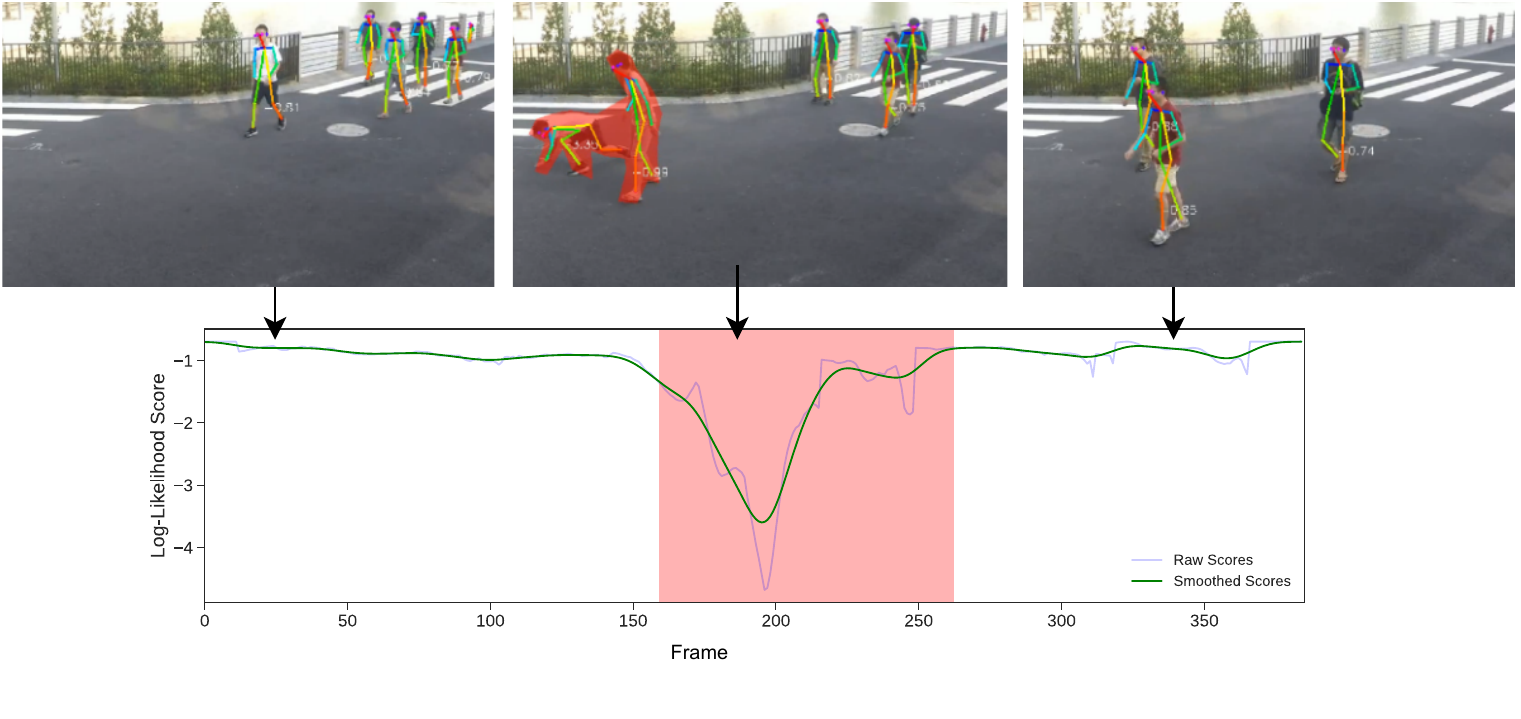}
        \caption{}
        \label{fig:UBnormal_a}
    \end{subfigure}
    \hfill
    \begin{subfigure}{\textwidth}
    \includegraphics[width=1.0\textwidth]{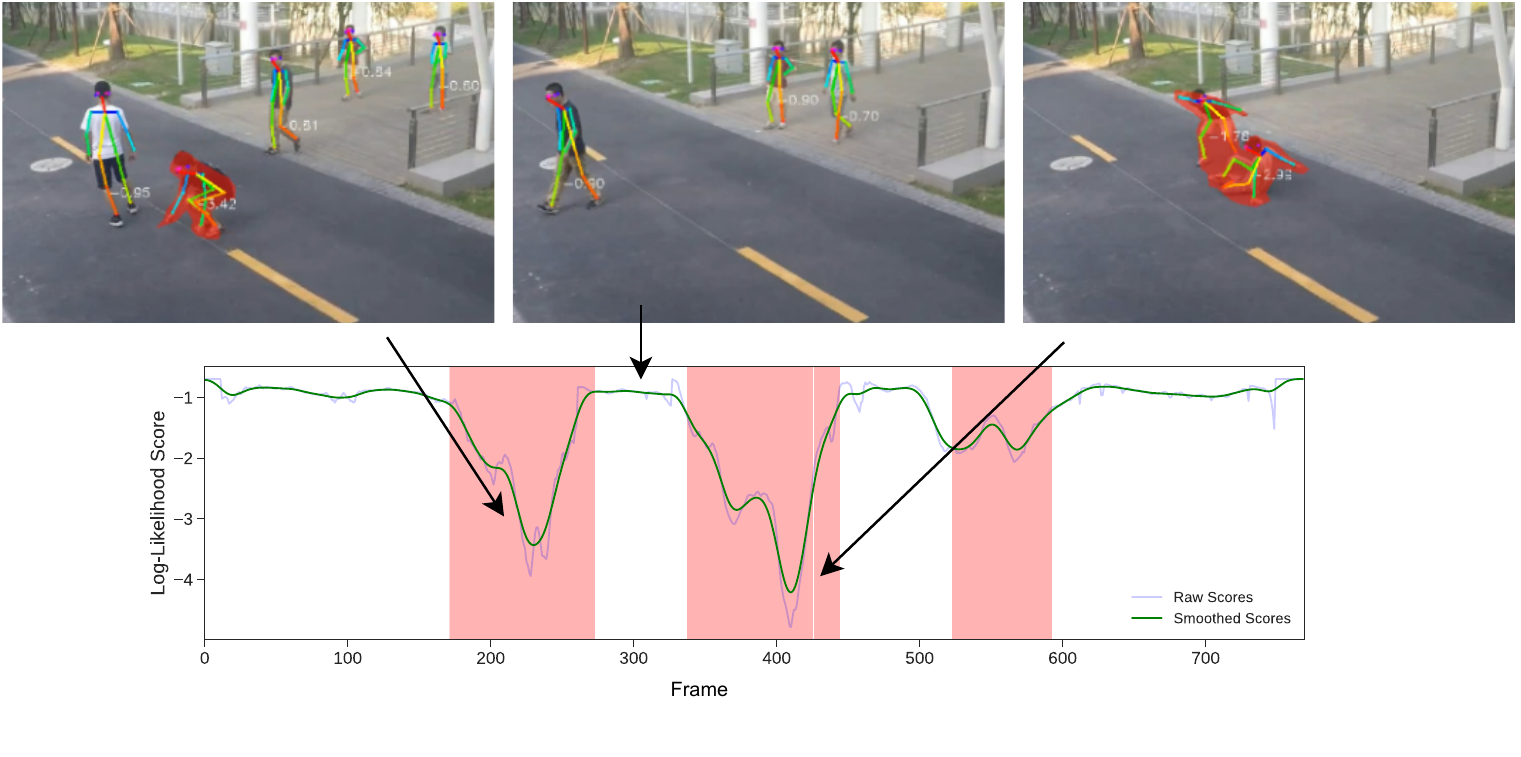}
        \caption{}
        \label{fig:UBnormal_a}
    \end{subfigure}
    \hfill
\caption{\textbf{Score examples for ShanghaTech dataset video}. Ground truth anomalous frames and people are marked red. Our unsupervised method is able to correctly recognize strange behavior and violence in both time and space.}
\label{fig:shanghaitech_5_23}
\end{figure*}

\begin{figure*}
\centering
    \begin{subfigure}{\textwidth}
    \includegraphics[width=1.0\textwidth]{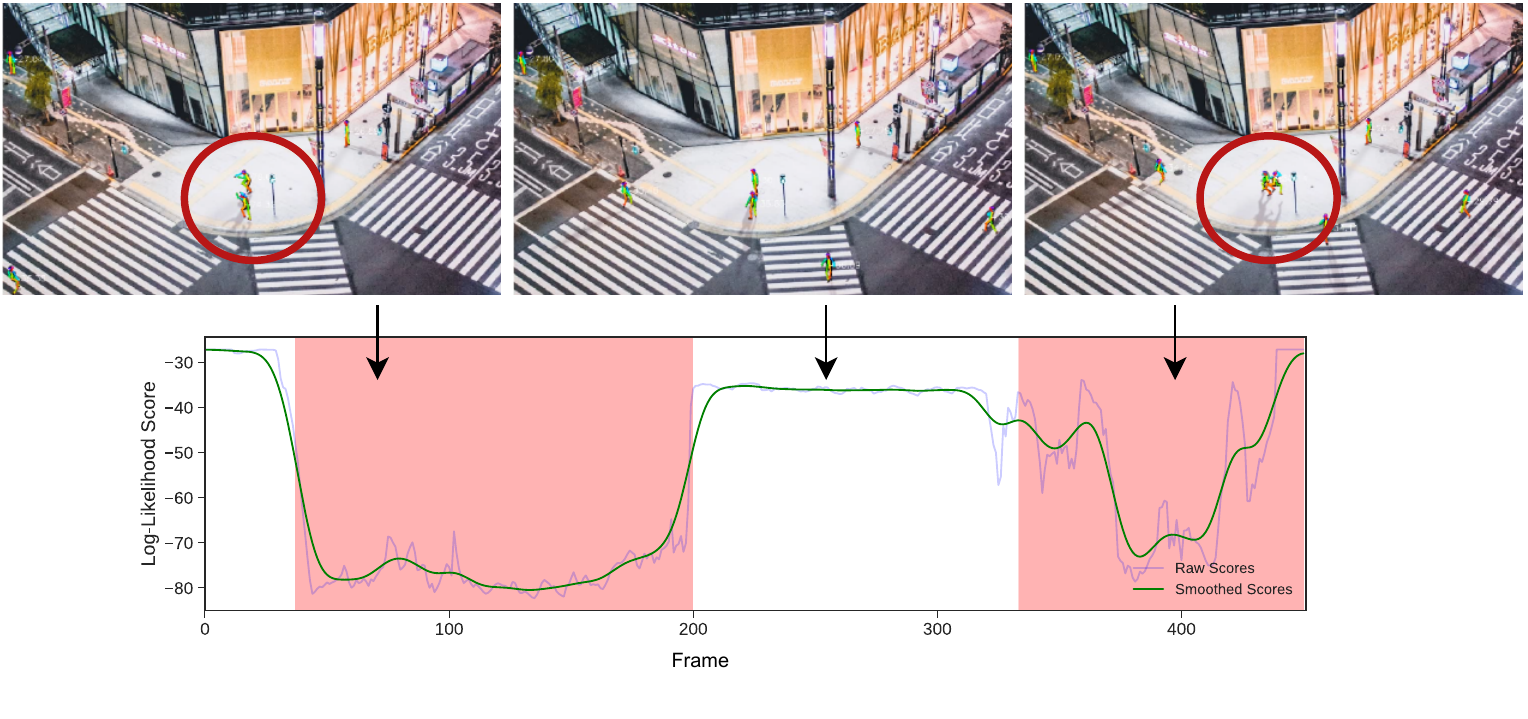}
        \caption{}
        \label{fig:UBnormal_a}
    \end{subfigure}
    \hfill
        \begin{subfigure}{\textwidth}
        \includegraphics[width=1.0\textwidth]{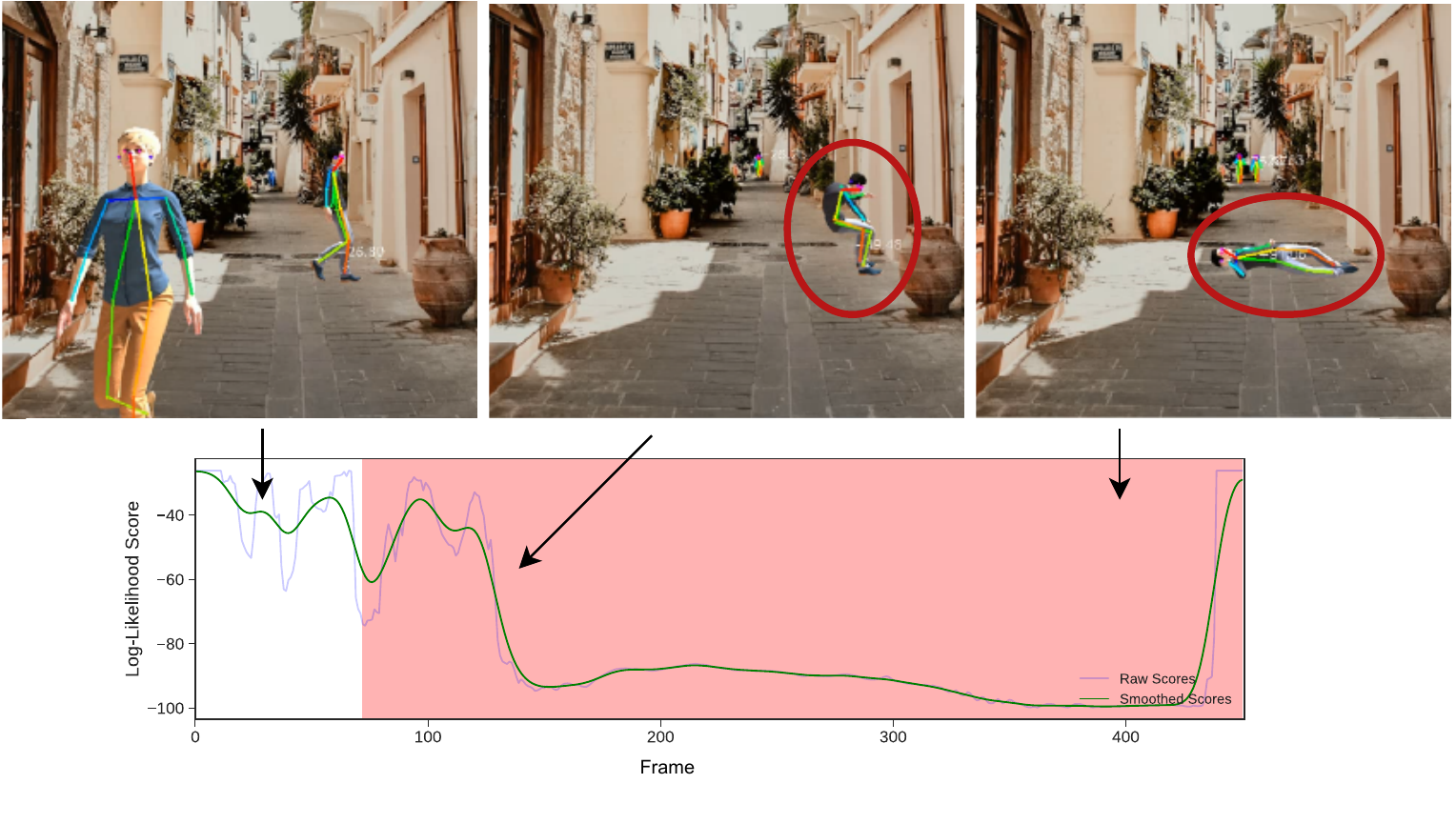}
        \caption{}
        \label{fig:UBnormal_a}
    \end{subfigure}
    \hfill
\caption{\textbf{Score examples for UBnormal dataset video}. Ground truth anomalous frames are marked red, and anomalies in photos are marked in a red circle.
(a) A scene of people dancing  - an abnormal behavior in the UBnormal dataset.
(b) A scene of a person falling to the ground.
Our supervised method is able to recognize anomalies even in difficult viewpoints correctly.}
\label{fig:ubnormal_17_4}
\end{figure*}

In this section, we showcase more frame-level AUC performance of our model. Figure \ref{fig:shanghaitech_5_23} shows violent behaviors in the ShanghaiTech datasets. Figure \ref{fig:ubnormal_17_4} shows people passing out or dancing (considered anomalous behavior) from the UBnormal dataset; our method successfully recognizes the anomalies despite the difficult viewpoint.

\label{results}
\end{document}